\def\year{2021}\relax

\documentclass[letterpaper]{article}
\usepackage{aaai21}
\usepackage{times}
\usepackage{helvet}
\usepackage{courier}
\usepackage[hyphens]{url}
\usepackage{graphicx}
\usepackage{graphicx}
\usepackage{natbib}
\usepackage{caption}
\usepackage{algorithm}
\usepackage[noend]{algorithmic}
\usepackage{diagbox}
\usepackage{subfig}
\usepackage{amsmath}
\usepackage{amsthm}
\usepackage{booktabs}

\newcommand{\agni}{\ensuremath{\text{Agni}}}

\title{Predicting Forest Fire Using \\ Remote Sensing Data And Machine Learning}
\author {
         Suwei Yang, \textsuperscript{\rm 1}
         Massimo Lupascu, \textsuperscript{\rm 2}
         Kuldeep S. Meel, \textsuperscript{\rm 1} \\
}
\affiliations {
     \textsuperscript{\rm 1} School of Computing, National University of Singapore \\
     \textsuperscript{\rm 2} Department of Geography, National University of Singapore \\
     \{suwei.yang, meel\}@comp.nus.edu.sg, \\
     mlupascu@nus.edu.sg
}

\begin{document}

\maketitle

\begin{abstract}
    Over the last few decades, deforestation and climate change have caused increasing number of forest fires. In Southeast Asia, Indonesia has been the most affected country by tropical peatland forest fires. These fires have a significant impact on the climate resulting in extensive health, social and economic issues. Existing forest fire prediction systems, such as the Canadian Forest Fire Danger Rating System, are based on handcrafted features and require installation and maintenance of expensive instruments on the ground, which can be a challenge for developing countries such as Indonesia. We propose a novel, cost-effective, machine-learning based approach that uses remote sensing data to predict forest fires in Indonesia. Our prediction model achieves more than 0.81 area under the receiver operator characteristic (ROC) curve, performing significantly better than the baseline approach which never exceeds 0.70 area under ROC curve on the same tasks. Our model's performance remained above 0.81 area under ROC curve even when evaluated with reduced data. The results support our claim that machine-learning based approaches can lead to reliable and cost-effective forest fire prediction systems.
\end{abstract}

\section{Introduction}

Over the last few decades, forest fires, especially peatland forest fires, have been a major environmental issue in Southeast Asia, with significant impacts on the atmosphere, carbon cycle and various ecosystem services \cite{Page2009}. Forest fires can affect a large portion of the population, causing economic difficulties and disruption to businesses and short and long term health problems \cite{doi:10.1111/gcb.12539,doi:10.1111/cobi.12662,Huijnen-2015-fire-carbon-emission,acp-16-11711-2016}. The smoke and haze produced by the fires can also cause disruptions and economic losses to neighbouring countries in hundreds of millions \cite{doi:10.1098/rstb.2015.0345,indonesia-forest-fire-cost}. Thus it is important to be able to predict forest fires and take mitigating measures to reduce their devastating impacts.

We focus on Indonesia because it accounts for 83\% of total peatland forests by area in Southeast Asia \cite{peat-swamp-forests-of-sea}. Peatland forests are important primarily because of their function as carbon storage \cite{doi:10.1111/j.1365-2486.2010.02279.x}. If peatland forests are burnt, the stored carbon would be released into the atmosphere as CO2, CH4, CO, etc. Indonesia is estimated to have 28.1 gigatons of carbon pool \cite{indonesia-peatland-carbon-estimate} in its peatlands. Whenever there is a large scale forest fire, the effects are devastating as shown by previous events of 1998, 2015 and 2019. For instance, in 2015 4.6M hectares were burned and 0.89 gigatons of carbon dioxide equivalent were released \cite{spatial-eval-indo-2015-fire-carbon}.

Existing forest fire prediction systems mainly take a modeling approach and use handcrafted features to perform predictions \cite{CFFDR,Farsite,EFFIS}. One existing system is the Canadian Forest Fire Danger Rating System \cite{CFFDR}, which is designed to evaluate environmental factors that influence the ignition, spread, and behavior of wildland fire. The system is managed by the Canadian government and has been adopted by other organizations around the world such as the New Zealand Forest Service \cite{nz-canada-forest-fire-danger-system} and European Commission which oversees the European Forest Fire Information System (EFFIS) \cite{europe-canadian-fwi-integration}.

Modeling-based approaches with handcrafted features and closed-form equations suffer from two drawbacks. Firstly, comprehensive fire prediction models have to deal with numerous weather-related and soil variables and their complex inter-relationships. It is difficult to describe all these intricate relationships with a set of mathematical equations. Consequently, scientists working on fire prediction systems approximate relationships between various weather-related factors by introducing simplifying assumptions \cite{KLOPROGGE2011289}. This might lead to a lack of robustness in the predictions, requiring system recalibration according to different deployment environments. Secondly, the models tend to rely on data from instruments on the ground and this leads to high operating costs. For example, the Canadian Forest Fire Danger Rating System operates on data from over 750 weather stations on the ground. Operating a similar system may be challenging in terms of cost for a developing country such as Indonesia. In light of the previously mentioned drawbacks coupled with recent advances in machine learning techniques and availability of remote sensing data, we want to answer \textit{Can machine-learning based techniques be used for forest fire prediction using remote sensing data?}

In this work, we use supervised learning to train a neural network with remote sensing data to predict forest fires in Indonesia. In particular, the remote sensing data consists of historical Landsat 7 satellite images \cite{landsat7-handbook} and Fire Information for Resource Management System (FIRMS) \cite{FIRMS-system} hotspot dataset. We designed our model to predict forest fires one month into the future, in line with the prediction time-frame of prior work \cite{spatial-lrl}. The time-frame requirement for prediction into the future stems from the need for time to plan resource intensive operations in remote terrains to either respond to potential forest fires~\cite{spatial-lrl} or prevent fires in the first place. More formally, at time $t$, our model takes in a year of historical data (i.e., sliding window of data belonging to  $  [t-52 \mathrm{week}, t]$) and outputs confidence of the presence of fire for $ t+5 \mathrm{week}$, which is referred to as {\em fifth week into future} for rest of the paper.

We performed extensive empirical evaluations on the prototype implementation of our system, called {\agni}. In particular, {\agni} achieves more than 0.81 area under the receiver operator characteristic (ROC) curve in our evaluations. We also performed comparative studies with the logistic regression-based baseline that was proposed in prior work \cite{spatial-lrl}. {\agni} performed significantly better than the baseline. The baseline approach never exceeded 0.70 area under ROC curve while {\agni} always stayed above 0.80.

Given the need for cost-effectiveness, we further evaluated {\agni} with reduced data. We limited the duration of historical data available to {\agni} for training and prediction to just 3 months and its performance still remained above 0.81 area under the ROC curve. In summary, our empirical evaluation demonstrates that {\agni}, a machine-learning based approach using remote sensing data, is effective while being more cost-effective than existing systems that required instruments on the ground. 

The rest of the paper is structured as follows. First, we discuss related prior works in Section \ref{sec:related-work}, then we provide details of the data used and its challenges in Section \ref{sec:data}. In Section \ref{sec:approach}, we discuss the details of our approach and the challenges encountered. Subsequently, we provide details of the evaluation in Section \ref{sec:results} and conclude in Section \ref{sec:conclusion}.

\section{Related work} \label{sec:related-work}

Existing forest fire forecasting systems, such as the popular Canadian Forest Fire Weather Index System \cite{CWFIS-1}, rely on data from weather stations (over 750 in this case) on the ground to operate. For the Canadian Forest Fire Weather Index System, the input data is then processed according to closed-form mathematical equations to attain Fire Weather Index (FWI) which is a numerical rating of fire intensity. As discussed above, our work seeks to develop alternative techniques based on publicly available satellite image data to provide cost-effective prediction systems for developing countries such as Indonesia.

While our work is the first in its kind to use neural network with remote sensing data to predict forest fires, a related study by Maulana et al. (\citeyear{spatial-lrl}) tried to predict peatland forest fires using a spatial logistic regression approach. Specifically, Maulana et al. employ careful expert-driven manual feature engineering involving features such as river network density and peat soil depth. Such features, while useful in their smaller case study, are not readily available across Indonesia. In contrast, our model takes only the satellite images as input, and is not restricted to peatland forest fires but predicts for forest fires in general.

Apart from the study by \citeauthor{spatial-lrl}, there are other existing machine-learning based approaches in the domain of forest fires but they do not specifically focus on fire prediction. Wijayanto et al. used Adaptive Neuro Fuzzy Interface System to predict whether a fire alarm trigger is a false positive and explored the relationship of fire with human activities \cite{ANFSI-1755-1315-54-1-012059}. Cortez and Morais used logistic regression, decision tree and support vector machines to predict the potential burn area given weather station data \cite{Cortez2007ADM}. FireNet performs real-time detection and annotation of fire boundaries in drone video feeds \cite{doshi2019firenet}. FireCast, on the other hand, tries to predict how a fire will spread in the near future \cite{ijcai2019-636}.

\section{Data} \label{sec:data}

Supervised learning requires both input data and ground truth labels to train a model. We used Landsat 7 satellite images, retrieved from Google Earth Engine \cite{gorelick2017google}, as inputs to the neural network. For ground truth labels, we used FIRMS hotspot dataset and we refer to forest fires as hotspots for consistency with FIRMS. In this section, we discuss details of the data used and their associated challenges. Finally, we describe the data preparation process.

\subsection{Landsat 7 satellite images}

Landsat 7 \cite{landsat7-handbook} satellite is part of the Landsat program, under NASA and U.S. Geological Survey. Landsat 7 has an orbit period of 16 days. Satellite images from Landsat 7 are used as the input source for our hotspot prediction model. The imaging instrument on Landsat 7 satellite, Enhanced Thematic Mapper Plus(ETM+), produces imagery of Earth as eight spectral bands. The satellite image specifications are as shown in Table \ref{tab:landsat7-spec}. 
Band column shows the respective names of the Landsat 7 satellite bands. Resolution column shows the resolution of the respective band. A resolution of 30m means that one pixel in the image band represents an area of $30m \times 30m$. Channel column shows what each respective band represents and Wavelength column shows the respective wavelengths of each band in $\mu m$.

\begin{table}[htb]
	\small
	\centering
	\begin{tabular}{l r r r} 
		\toprule
		Band & Resolution & Channel & Wavelength $(\mu m)$ \\
		\midrule
		B1 & 30m & Blue  & 0.45-0.52\\ 
		B2 & 30m & Green & 0.52-0.60\\
		B3 & 30m & Red & 0.63-0.69\\
		B4 & 30m & Near IR & 0.77-0.90\\
		B5 & 30m & IR 1 Shortwave & 1.55-1.75\\
		B6{\_}1 & 60m & Low-gain IR 1& 10.40-12.50\\
		B6{\_}2 & 60m & High-gain  IR 1 & 10.40-12.50\\
		B7 & 30m & Shortwave  IR 2 & 2.09-2.35\\
		B8 & 15m & Panchromatic & 0.52-0.90 \\
		\bottomrule
	\end{tabular}
	\caption{Specifications of different bands on Landsat 7}
	\label{tab:landsat7-spec}
\end{table}

Six of the bands have a resolution of 30m, except bands B6 and B8. However, band B6 is already resampled to the resolution of 30m in the Landsat 7 data available on Google Earth Engine. Since all bands except band B8 have the same resolution ($30m \times 30m$), we focus on usage of bands B1 to B7.  

\subsection{Fire Information for Resource Management System (FIRMS) hotspot data}

In order to train a neural network, we need to provide ground truth labels of fire. We found the FIRMS hotspot dataset \cite{FIRMS-system} to be the most comprehensive open fire dataset available. The FIRMS dataset, provided by NASA, is used by several fire prediction systems such as the Canadian Wildland Fire Information System, the European Forest Fire Information System and the U.S. Forest Service - Missoula Fire Sciences Lab. In this work, we use the FIRMS hotspot data as labels for training and evaluation of prediction models.

The FIRMS dataset consists of hotspots detected by LANCE \footnote{Land, Atmosphere Near real-time Capability for EOS} system after processing the MODIS \footnote{Moderate Resolution Imaging Spectroradiometer} and VIIRS \footnote{Visible Infrared Imaging Radiometer Suite} instrument data on Terra, Aqua EOS and Suomi-NPP satellite \cite{lance,modis,viirs,terra,aqua,suomi-npp}. The resolution of the FIRMS dataset is 1km $\times$ 1km per hotspot pixel. Each detected hotspot instance is associated with a confidence level ranging from 0\% to 100\%, indicating certainty of the detection. For a given area, we assign the label to be the maximum confidence value within that area.

\subsection{Challenges} \label{sec:data-challenges}

\subsubsection{Imperfect orbit and damaged SLC}

\begin{figure}[htp]
	\small
	\centering
	\includegraphics[width=0.7\columnwidth]{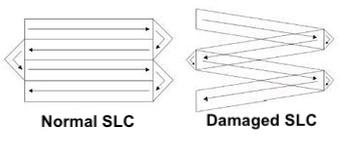}
	\caption{Effect of damaged SLC (source: U.S. Geological Survey)}
	\label{fig:slc-off}
\end{figure}

We encountered two primary sources of missing data: imperfect orbit of satellite and damaged scan-line calibrator (SLC). The first source of missing data arises from Landsat 7's imperfect orbit. Ideally, a satellite with a perfect orbit can produce images of the exact same area during each orbit. In reality, the Landsat 7 satellite images do not overlap perfectly with each other between orbits and sometimes parts of an area are missing due to imperfect orbit. To overcome the issue of imperfect image overlap, we choose a reference image and crop the remaining satellite images to the area of the reference image. When the satellite images are sliced up into smaller images, to the correct prediction area, some slices may be missing, i.e. entire image slice is filled with zero entries.

The other source of missing data mentioned earlier is Landsat 7's scan-line calibrator (SLC). The scan-line calibrator (SLC) ensures that the imaging instrument on Landsat 7 satellite scans in the correct manner when producing satellite images in a row-wise manner. A normal imaging motion with a working SLC is shown by the arrows on the left of Figure \ref{fig:slc-off}. When the SLC is damaged in the case of Landsat 7, the imaging motion is in a zigzag form, shown by the arrows on the right of Figure \ref{fig:slc-off}. The resulting satellite image produced by the damaged SLC has triangular strips of missing data.

For both sources of missing data encountered, we did not perform statistical imputation.\footnote{Statistical imputation refers to the process of replacing missing data through statistical techniques.} Our hypothesis is that the model should be able to learn from the remaining data. Our results show that the model is still able to perform well, supporting our hypothesis of the models learning ability on the remaining data. More details will be discussed in Section \ref{sec:results}.

\subsubsection{Resolution and data size trade-off}

Ideally, we would want to use satellite image data at full resolution (30m by 30m per pixel for Landsat 7) for model training and evaluation. However, the capacity to handle full resolution images requires extensive storage. At full resolution, a reference satellite image area and its associated historical satellite images require around 2 gigabytes of storage. We used 800 such satellite image time series, requiring more than 1.6 terabytes of storage if full resolution satellite images are used. File size is still an issue, even when the image is downscaled. As a result, each image in a historical satellite image time series is first downscaled, then sliced into smaller images and subsequently converted to a histogram. In particular, each band of the image is converted to a histogram of 32 evenly spaced intervals that are referred to as bins, with each bin representing counts of 8 pixel values as each band has pixel value range from 0 to 255. For an image band $b$, the first bin in the histogram would be the total count of pixel values in band $b$, in the range of 0 to 7 inclusive. Since we work with 8 bands (B1 to B7), each image is converted into 8 histograms of 32 bins. The resultant histograms are a storage efficient representation of an 8km by 8km area. 

\subsection{Data preparation}

\begin{figure}
	\small
	\centering
	\includegraphics[width=0.9\columnwidth]{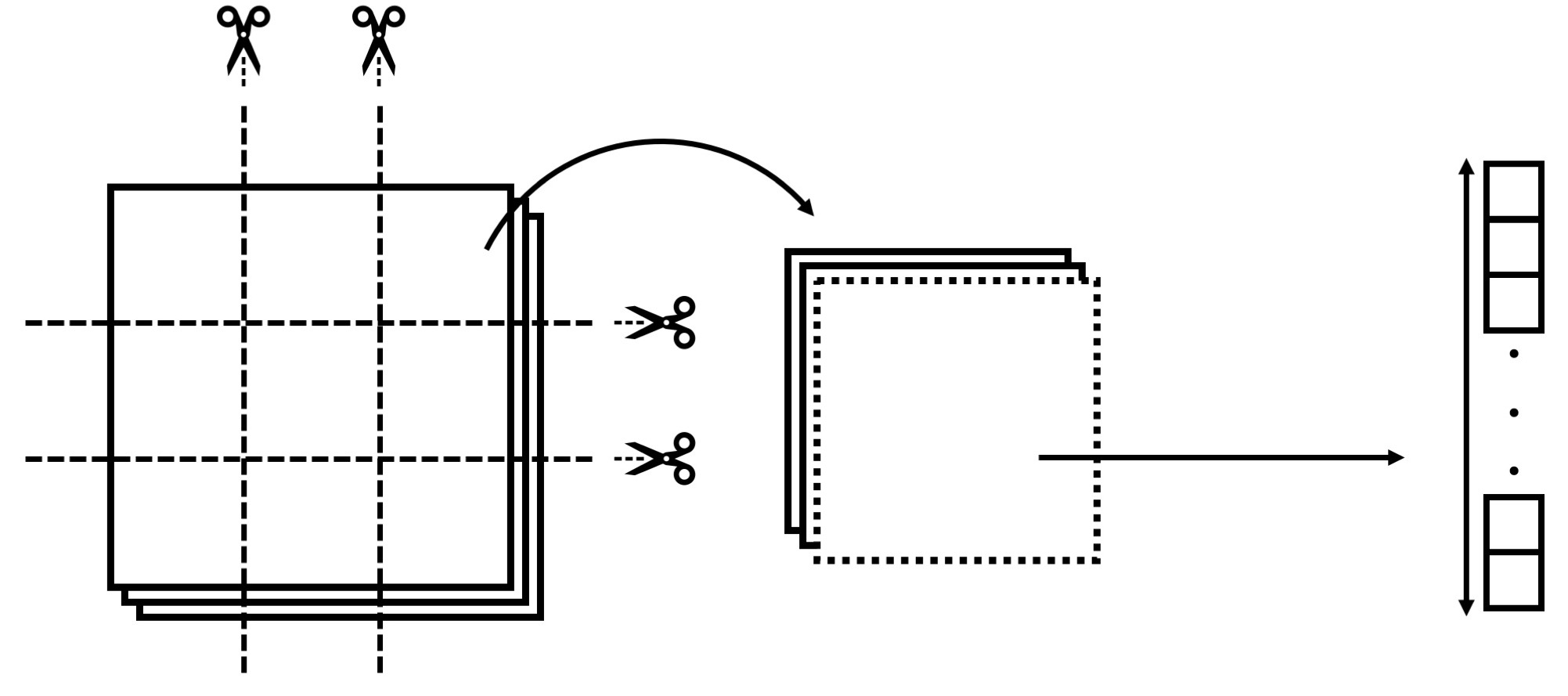}
	\caption{Data preparation process: The big satellite image time series is sliced into 8 x 8km area time series, then each channel is converted to histogram of 32 equal bins}
	\label{fig:data-preproc}
\end{figure}

\begin{algorithm}[tb]
	\begin{flushleft}
	\textbf{Input}: $t_{s}$ and $t_{e}$ which represents start and end date\\
	\textbf{Output}: Historical satellite image histogram and future labels (5th week into future)
	\end{flushleft}
	\begin{algorithmic}[1]
		\STATE refImgList $\longleftarrow$ getRefImgs($t_{s}$, $t_{e}$) \label{line:preproc:getRefIndoImg}
		\FOR{\text{$r_{i}$ in refImgList}}
		\STATE $t_{r_{i}}$ $\longleftarrow$ getDate($r_{i}$) \label{line:preproc:refDate}
		\STATE pastImgs $\longleftarrow$ getHistoricalImgs($r_{i}$, $52wk$) \label{line:preproc:getHistoricalData}
		\STATE label $\longleftarrow$ getFirmsLabel($r_{i}$, $t_{r_{i}} + 4wk$, $t_{r_{i}} + 5wk$) \label{line:getLabels}
		\STATE imgLabelTensor $\longleftarrow$ combine(pastImgs, label) \label{line:preproc:combineLabelAndData}
		\STATE slicedDataList $\longleftarrow$ slice(imgLabelTensor, $8km$) \label{line:preproc:sliceDataInst}
		\FOR{slicedData in slicedDataList} \label{line:preproc:toHistStart}
		\IF{!allZero(slicedData)} \label{line:ignoreZero}
		\STATE finalData $\longleftarrow$ toHist(slicedData) \label{line:getSlicedFinalData}
		\STATE save(finalData)
		\ENDIF
		\ENDFOR \label{line:preproc:toHistEnd}
		\ENDFOR
	\end{algorithmic}
	\caption{dataPrep algorithm: algorithm to retrieve and preprocess satellite images and FIRMS labels. $wk$ stands for week.}
	\label{alg:preprocess-algo}
\end{algorithm}

In the data preparation stage, retrieved historical satellite images are sliced into smaller images that represent 8km by 8km areas. Subsequently, the sliced images are then converted to histograms and paired with the respective ground truth hotspot label from the FIRMS data. The data preparation process is illustrated in Figure \ref{fig:data-preproc} and Algorithm \ref{alg:preprocess-algo} provides further details about the process.
 
In line \ref{line:preproc:getRefIndoImg}, the algorithm retrieves all reference satellite images of Indonesia within the interval specified by the input $t_s$ and $t_e$. For each reference image, the past one-year of satellite images of the same area are retrieved in subroutine \textit{getHistoricalImgs} on line  \ref{line:preproc:getHistoricalData}. The \textit{getHistoricalImgs} subroutine crops historical images to the exact area of the reference image. The \textit{getFirmsLabel} subroutine on line \ref{line:getLabels} retrieves FIRMS hotspot data for the fifth week in the future with respect to the reference image date. The highest FIRMS value per pixel within the week long hotspot data is taken as the ground truth label. Line \ref{line:preproc:combineLabelAndData} combines the label with corresponding historical images. The combined image and label data are then sliced into smaller pieces to achieve the correct area of 8km by 8km in line \ref{line:preproc:sliceDataInst}. If an image is not entirely filled with zeros due to missing data (line \ref{line:ignoreZero}), the subroutine \textit{toHist} on line \ref{line:getSlicedFinalData} converts each band of the sliced data into a histogram of 32 bins.

\section{Approach and Challenges} \label{sec:approach}

\begin{figure*}[!ht]
	\centering
	\subfloat[September 2019 hotspots, AUC = 0.8480]{\includegraphics[width=0.35\textwidth]{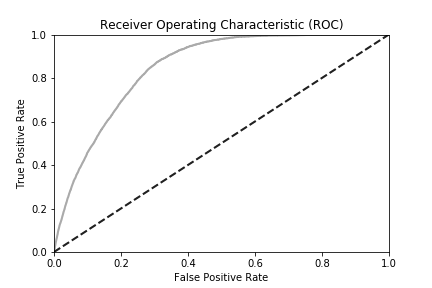} \label{fig:rocaug19}}
	\hspace{0.8cm}
	\subfloat[August 2019 hotspots, AUC = 0.8428]{\includegraphics[width=0.35\textwidth]{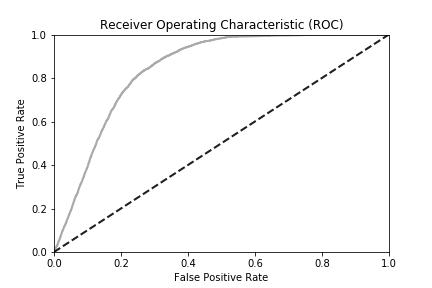}\label{fig:rocjuly19}}
	\vspace{0.01cm}
	\caption{Area under ROC curve(AUC) for evaluation on (a)September and (b)August 2019 hotspots} \label{fig:roc-curves-both}
\end{figure*}

We employed neural networks for the supervised learning task. We first present the architecture of the neural network employed and then shed light on two key design choices to handle the challenges specific to our application domain.  

\subsection{Neural network architecture}

\begin{table}[htp]
	\centering
	\begin{tabular}{l r r r} 
		\toprule
		Layer & Operation & Hidden Units\\
		\midrule
		1 & Conv2D (3 $\times$ 3) & 64\\ 
		2 & LSTM  & 64\\
		3 & Dropout & -\\
		4 & Dense (ReLU) & 256\\
		5 & Dense (ReLU) & 32\\
		Output & Dense (Sigmoid) & 1\\
		\bottomrule
	\end{tabular}
	\caption{Neural network architecture}
	\label{tab:nn-architecture}
\end{table}

The neural network architecture is shown in Table \ref{tab:nn-architecture}. The Layer column indicates the positions of layers in the architecture. The Operation column indicates the type of layer and the Hidden Units column indicates the number of hidden units in the respective layer. The neural network takes as input a histogram time series that represents an 8km by 8km area. The input is a one year time series of satellite image histograms with $T$ timesteps. Each timestep consists of information from 8 bands, with 32 equally divided histogram bins for each band. The network first performs convolution \cite{Fukushima1980} on each input timestep of the data, which is the histogram of individual satellite images in the one year of historical satellite images. The convolution with 64 kernels of size 3 $\times$ 3 is applied to histogram bins of the different bands to learn representations of relationships between the different bands. The representation attained from the convolution layer, during each timestep, is passed to the Long-Short Term Memory (LSTM) layer \cite{lstm} with 64 hidden units. The LSTM layer learns the temporal relationship of the historical data. This is followed by dropout with a rate of 0.3 and three dense layers with 256, 32 and 1 hidden units respectively. The last dense layer uses \textit{sigmoid} activation to get an output value between 0 and 1. The output is interpreted as the predicted confidence of hotspot in that area for the fifth week into the future, from the most recent historical data timestep.

The neural network has a custom loss function $L$ as follows:

\begin{align*}
L = & | y_{true} - y_{pred}| \times \\
& \qquad \qquad min(10^{5}, max(1, 10^{( (y_{true} - y_{pred}) \times \frac{100}{k})}))
\end{align*} \label{eqn:custom-loss}

$y_{true}$ is the label and $y_{pred}$ is the predicted value for confidence of hotspot presence. Both $y_{true}$ and $y_{pred}$ take values between 0 and 1. In $L$, $k$ is a hyper-parameter set to be 30 for best performance after hyper-parameter tuning. Other values of $k$ that we have tried include 100, 50 and 10.

The choice of loss function is primarily influenced by two factors: (1) the presence of skewed data, and (2) the cost of mis-predictions. As discussed below, the input data is heavily skewed towards absence of hotspots (i.e., $y_{true}$ being mostly 0). The presence of skewed data requires us to steer the model away from always predicting $y_{pred}$ to be close to 0, and therefore, we provide high penalty for the case of $y_{true}> y_{pred}$. Secondly, in practice, higher values of $y_{pred}$ may give rise to false alarms but lower values of $y_{pred}$ than the actual ground truth leads to unpreparedness and as such, leads to substantial loss of human, social, and economic capital.  

\subsection{Challenges} \label{sec:challenges-approach}

\subsubsection{Heavily skewed data}

As briefly mentioned above, the natural distribution of the data is heavily skewed towards absence of hotspots. Empirical analysis of the data indicates that for a typical month, over 99\% of locations do not have presence of hotspots. As such, a machine learning model may achieve 99\% accuracy by simply predicting $y_{pred}$ to be 0. In addition to the design of the custom loss function $L$ discussed above, we perform the standard practice of resampling of the training data to achieve a 3:1 ratio for zero to non-zero labels. 

\subsubsection{Batch size limitations during training}
As discussed in Section~\ref{sec:data-challenges}, the imperfect orbit of Landsat 7 leads to missing data for several timesteps when the data is viewed as a time series. The lack of data for certain timesteps has implications on the batch size for the training procedure of the neural network. In order to perform batch training, our data has to be of the same dimensions. However, that is not the case for our data because of missing timesteps in our data. Therefore, we train the neural network with a batch size of 1. An alternate approach would have been the usage of padding to ensure that the input training data instances within the same batch have the same dimensions. The lack of prior research about possible padding approaches for satellite image time series limited its use in our work but we believe exploring its role in future work is an interesting technical challenge.

\section{Evaluation setup and results} \label{sec:results}

\begin{table*}[h!]
	\small
	\centering
	\begin{tabular}{l r r r r} 
		\toprule
		\diagbox{Hotspot\\ label}{Window} & $\sim$3 Month & $\sim$6 Month & $\sim$9 Month & 1 Year\\
		\midrule
		June'19 & 0.8344 & 0.8428 & 0.8277 & 0.8125 \\
		July'19 & 0.8623 & 0.8707 & 0.8596 & 0.8519 \\
		Aug'19 & 0.8269 & 0.8471 & 0.8511 & 0.8428 \\
		Sept'19 & 0.8239 & 0.8464 & 0.8449 & 0.8480 \\ 
		\bottomrule
	\end{tabular}
	\caption{Area under ROC curve values for varying size of sliding window}
	\label{tab:roc-train-months}
\end{table*}

\begin{table*}[h!]
	\centering
		\begin{tabular}{l r r r r} 
		\toprule
		\diagbox{Hotspot\\ label}{Model} & Agni & Agni(MSE) & LR(custom loss) & LR(MSE)\\
		\midrule
		June'19 & 0.8125 & 0.7483 & 0.6966 & 0.6355\\
		July'19 & 0.8519 & 0.7081 & 0.7312 & 0.6967\\
		Aug'19 & 0.8428 & 0.7264 & 0.7171 & 0.6613 \\
		Sept'19 & 0.8480 & 0.6956 & 0.7304 & 0.6560 \\
		\bottomrule
		\end{tabular}
	\caption{LR stands for logistic regression baseline and MSE stands for mean squared error loss function}
	\label{tab:custom-loss-ablation}
\end{table*}

To empirically validate our models, we implemented the prototype, {\agni}, using Tensorflow 1.12 and Python 2.7. We conducted the training and evaluation on high performance machines on compute clusters with the following specifications: Intel Xeon Silver 4108 with Nvidia Titan V and Intel Xeon Silver 4108 with Nvidia Tesla V100.
 
We employ a logistic regression approach as baseline, similar to prior work by \citeauthor{spatial-lrl}. In our baseline, we used mean squared error loss instead of binary cross entropy loss as the labels are real numbers from 0 to 1. 

The choice of training dataset is influenced by three key considerations:  (1) the need to minimize the amount of training data for computational efficiency, (2) the usage of data with less heavy skew towards the absence of hotspots, and (3) the usage of recent data so as to allow testing and validation of the model for recent months. Therefore, we employed data from 2014-18 as they coincided with large scale forest fires in Indonesia. We employed hotspots from April'18 as test set and hotspots from June'19 to September'19 as our validation set. Recall that we train {\agni} to predict labels based on historical data for the past year. In particular, the model is trained to predict labels for the 5th week ($t+5 \mathrm{week}$) given a sliding window of one year of data, i.e., from $[ t-52 \mathrm{week}, t]$. This prediction period of 5th week into future is also in line with the spatial logistic regression work \cite{spatial-lrl}. As mentioned in Section \ref{sec:challenges-approach} the training data is of 3:1 ratio for zero to non-zero labels. Apart from 3:1 ratio, we also experimented with other ratios, specifically 4:1 and 2:1, but did not see a significant change in {\agni}'s performance.

The objective of our evaluations was to answer the following questions:
\begin{description}
	\item[RQ 1] How well does {\agni} predict hotspots? 
	\item[RQ 2] How does the length of sliding window affect {\agni}'s performance?
	\item[RQ 3] How does our custom loss function affect model performance? 
\end{description}

\subsection{Performance of Model (RQ 1)}

For unambiguous communication with general public and across different agencies, we are often interested in the binarized output: whether there is going to be a hotspot or not. The standard process in machine learning community is to choose a threshold $\tau$ and binarize the output value based on $\tau$. Any prediction value greater than $\tau$ is classified as \textit{hotspot}, otherwise the prediction is classified as \textit{no hotspot}. We interpret a non-zero label from FIRMS as \textit{hotspot} and zero label from FIRMS as \textit{no hotspot}. We plot the Receiver Operating Characteristic (ROC) curve\footnote{The relevant notation definitions are included in the appendix}, Figure \ref{fig:roc-curves-both}, to show the behavior of our model across different $\tau$ values and we use area under ROC curve (AUC) as an indication of the model's performance. Figure \ref{fig:roc-curves-both} shows the ROC curves for August and September'19 and the AUC values for both August and September'19 are above 0.84. In this context, it is worth noting that if a model were to always predict no fire, the expected AUC would be 0.5, illustrated by the dotted lines in Figures \ref{fig:rocaug19} and \ref{fig:rocjuly19}. 

\begin{table}[h!]
	\small
	\centering
	\begin{tabular}{l r r} 
	\toprule
	Hotspot label & Total labels & Positive hotspot labels\\
	\midrule
	June'19 & 122416 & 495 (0.4\%)\\
	July'19 & 95594 & 1017 (1.1\%)\\
	Aug'19 & 104725 & 2885 (2.8\%)\\
	Sept'19 & 115934 & 6216 (5.4\%)\\ 
	\bottomrule
	\end{tabular}
	\caption{Distribution of labels of evaluation data. Postive hotspot labels - labels indicating non-zero confidence of hotspot presence.  Total labels - total number of labels.}
	\label{tab:hotspot-dist}
\end{table}

Further evaluations of the prediction model are conducted with June and July hotspots in 2019. The model had an AUC of 0.8519 and 0.8125 for prediction of July'19 and June'19 hotspots respectively. As evident from Table \ref{tab:hotspot-dist}, there were significantly fewer hotspots for June and July'19 in comparison to August and September'19.  The consistent performance of {\agni} across different months with varying distribution of data provides strong evidence in favor of robustness and versatility of the model. It is also worth noting that our model performs significantly better than the logistic regression baseline as shown in Table \ref{tab:baseline-comparison}. In the evaluations, {\agni}'s AUC remained above 0.80 whereas the baseline's AUC never exceeded 0.70.

\begin{table}[h!]
	\small
	\centering
	\begin{tabular}{l r r} 
		\toprule
		Hotspot label & {\agni} (our model) & LR baseline\\
		\midrule
		June'19 & 0.8125 & 0.6355\\
		July'19 & 0.8519 & 0.6967\\
		Aug'19 & 0.8428 & 0.6613\\
		Sept'19 & 0.8480 & 0.6560\\ 
		\bottomrule
	\end{tabular}
	\caption{Area under ROC curve for Agni and LR baseline(logistic regression)}
	\label{tab:baseline-comparison}
\end{table}

\subsection{Impact of Sliding Window on Performance (RQ 2)}

To understand the data requirements of {\agni}, we train and evaluate {\agni} with the length of sliding window varying from 3 months to 1 year. Table \ref{tab:roc-train-months} shows the AUC values for the different lengths of sliding window used for training and prediction. Each column in Table \ref{tab:roc-train-months} indicates the length of sliding window and each row corresponds to the month of hotspots predicted. For example, the cell corresponding to Sept'19 and $\sim$3 Month shows the AUC value for {\agni} trained using a sliding window of the 6 most recent satellite images available and evaluated for Sept'19 hotspots.

\subsection{Impact of custom loss function on performance (RQ 3)}

To understand how much our custom loss function defined in Section \ref{eqn:custom-loss} contributes to model prediction performance for $t+5week$, we conducted ablation studies for the choice of loss function. We use the same evaluation data as before and compare against the performance of models in Table \ref{tab:baseline-comparison}. Specifically, we conducted additional evaluations on a verison of {\agni} with mean squared error as loss function and separately for a version of logistic regression model with our custom loss function. Our results, shown in Table \ref{tab:custom-loss-ablation}, show that our custom loss function greatly improves the performance of both models and establishes its necessity. When we trained logistic regression using our custom loss function, its performance could exceed 0.70 AUC unlike before. Similarly, when we trained {\agni} using mean squared error, its performance dropped to lower than 0.75 AUC and some cases worse than logistic regression trained with our custom loss function. This strongly supports our choice of custom loss function. 

\subsection{Summary}

In summary, {\agni} trained with our custom loss function performs consistently better than the baseline model across different evaluation scenarios, as demonstrated by Tables \ref{tab:baseline-comparison} and \ref{tab:roc-train-months}, thereby providing strong evidence in support of the predictive power of machine-learning techniques using remote sensing data. Furthermore, {\agni} can be trained and evaluated with a significantly smaller sliding window, thereby requiring less storage of data during deployment and thus positioning itself as a cost-effective fire prediction system for developing countries such as Indonesia. 

\section{Conclusion} \label{sec:conclusion}

In this work, we trained a neural network to perform forest fire prediction. While existing prediction models are effective, they use data collected from instruments on the ground, leading to high operating costs. Our model uses readily available remote sensing data in the form of satellite images to predict fire hotspots. We showed that our model performs consistently well under extensive evaluations, providing an affirmative answer to the initial question \textit{Can machine learning-based techniques be used for forest fire prediction using remote sensing data?}

An interesting area to explore in the future is using multiple data sources, such as precipitation, soil moisture and human activity data, to achieve better performance. In particular, human activity data may be useful for predicting fires started by humans. It is a common practice to use fire to clear land before planting crops, but the fire could spread to forest areas, becoming forest fires. The challenge in using multiple data sources would be integrating sources with different resolutions. This would require a redesigning of the neural network architecture to include multiple inputs and merging them within the network.

\section*{Acknowledgments}
This work was supported in part by National Research Foundation Singapore under its NRF Fellowship Programme[NRF-NRFFAI1-2019-0004] and AI Singapore Programme [AISG-RP-2018-005], and NUS ODPRT Grant [R-252-000-685-13]. This work uses some resources of the National Supercomputing Centre, Singapore (https://www.nscc.sg). We would also like to acknowledge Dr. Liew Soo Chin from Centre for Remote Imaging, Sensing and Processing (CRISP) and Mr. Kim Stengert from World Wide Fund for Nature (WWF) Singapore for insightful discussions.

\begin{small}
\bibliography{references}
\end{small}

\end{document}